\documentclass[sigconf]{acmart}
\AtBeginDocument{%
  }

\copyrightyear{2026}
\acmYear{2026}
\setcopyright{cc}
\setcctype{by}
\acmConference[CIKM '26]{Proceedings of the 35th ACM International Conference on Information and Knowledge Management}{November 07--11, 2026}{Rome, Italy}
\acmBooktitle{Proceedings of the 35th ACM International Conference on Information and Knowledge Management (CIKM '26), November 07--11, 2026, Rome, Italy}
\acmDOI{10.1145/3799682.3840036}
\acmISBN{979-8-4007-2539-5/2026/11}

\usepackage{hyperref}
\usepackage{multirow}
\usepackage{subcaption}
\usepackage{tcolorbox}
\usepackage{tcolorbox}
\usepackage{adjustbox}
\begin{document}

\title{Do General NLP Embeddings Capture Ontological Reasoning? }


\author{Hamed Babaei Giglou}
\orcid{0000-0003-3758-1454}
\email{hamed.babaei@tib.eu}
\affiliation{%
  \institution{TIB Leibniz Information Centre for Science and Technology}
  \city{Hannover}
  \state{Lower Saxony}
  \country{Germany}
}

\author{Jennifer D'Souza}
\orcid{0000-0002-6616-9509}
\email{jennifer.dsouza@tib.eu}
\affiliation{%
  \institution{TIB Leibniz Information Centre for Science and Technology}
  \city{Hannover}
  \state{Lower Saxony}
  \country{Germany}
}

\author{Sören Auer}
\orcid{0000-0002-0698-2864}
\email{auer@tib.eu}
\affiliation{%
  \institution{TIB Leibniz Information Centre for Science and Technology}
  \institution{L3S Research Center, Leibniz University of Hannover}
  \city{Hannover}
  \state{Lower Saxony}
  \country{Germany}
}

\renewcommand{\shortauthors}{Babaei Giglou et al.}

\begin{abstract}
General-purpose NLP embedding models perform well on linguistic tasks, but their ability to capture symbolic ontological structure remains unclear. We introduce AVA, a systematic framework for evaluating whether embeddings distinguish logic-sensitive relational semantics in ontologies and knowledge graphs. AVA comprises 171,007 contrastive triplets derived from 163 heterogeneous ontologies using hierarchy inversion, relation substitution, and disjointness injection. Each triplet contains an ontology statement, a semantically equivalent paraphrase, and a logic-sensitive hard negative with contradictory relational meaning. We evaluate more than 25 state-of-the-art embedding models and find substantial limitations: the best model achieves only 0.739 triplet accuracy, while hard negative accuracy falls to 0.135. Fine-tuning improves discrimination by a large margin but transfers poorly to downstream Semantic Web tasks, including taxonomy discovery and ontology alignment. Further analysis suggests that improvements stem partly from perturbation-specific pattern recognition rather than robust ontological understanding. These findings reveal a persistent gap between linguistic representation learning and ontology-level discrimination, challenging the assumption that strong NLP benchmark performance translates to Semantic Web competence.
\end{abstract}



\received{20 February 2007}
\received[revised]{12 March 2009}
\received[accepted]{5 June 2009}

\maketitle

\section{Introduction}

Recent advances in representation learning have transformed both Natural Language Processing (NLP) and the Semantic Web. General-purpose embedding models, ranging from Word2Vec~\cite{Word2Vec} and GloVe~\cite{Glove} to transformer-based architectures such as BERT~\cite{devlin2019bert}, MPNet~\cite{MPNet}, E5~\cite{wang2022text}, GTE~\cite{zhang2024mgte,li2023towards}, and BGE~\cite{bgeembedding}, achieve strong performance on retrieval and Semantic Textual Similarity (STS) benchmarks. In parallel, the Semantic Web community has developed ontology and knowledge graph embedding approaches, including RDF2Vec~\cite{ristoski2016rdf2vec}, OWL2Vec*~\cite{chen2021owl2vec}, DL2Vec~\cite{DL2Vec}, and knowledge graph embedding (KGE) methods such as TransE~\cite{TransE}, ComplEx~\cite{ComplEx}, and RotatE~\cite{RotatE}. More recent approaches, including KG-BERT~\cite{yao2019kg} and KEPLER~\cite{wang2021kepler}, attempt to combine linguistic representations with structured knowledge. Despite these advances, a fundamental challenge remains unresolved: \emph{generalization in Semantic Web representation learning}. Traditional KGE methods operate on fixed entity and relation vocabularies and often require retraining when applied to new ontologies~\cite{chen2023generalizing,survey2023}. Ontology-aware methods capture rich structural semantics within a particular ontology but may exhibit limited transferability across heterogeneous schemas~\cite{qiang2023ontology,bian2025llm,li2025towards}. Conversely, transformer-based sentence embeddings generalize well across linguistic tasks yet are not explicitly optimized for symbolic reasoning over OWL/RDFS structures. Hyperbolic representations provide a promising alternative for hierarchical knowledge~\cite{nickel2017poincare,dhingra2018embedding}, but their effectiveness for transferable ontology relation discrimination remains unclear.

This challenge is closely related to STS, a primary evaluation paradigm for sentence embeddings. Modern embedding models achieve strong STS performance~\cite{kumar2025advances}, suggesting they can capture semantic equivalence between textual expressions. However, standard STS benchmarks rarely require distinguishing between statements that are lexically similar yet differ in ontology-level relational semantics~\cite{gatto-etal-2023-text,sun2025text}. As a result, it remains unclear whether strong STS performance reflects genuine sensitivity to subclass relations, domain/range constraints, disjointness axioms, and other forms of symbolic knowledge. To investigate this question, we introduce \textbf{AVA}, a logic-sensitive ontology similarity benchmark based on structured ontology perturbations. AVA generates ontology-aware hard negatives through hierarchy inversion, relation substitution, and disjointness injection, producing sentence pairs that preserve substantial lexical overlap while expressing contradictory ontological meaning. Using 163 heterogeneous ontologies, we construct a dataset of 171,007 contrastive triplets and evaluate both pre-trained and fine-tuned embedding models under cross-ontology generalization settings.

Our experiments reveal three key findings. First, even the strongest general-purpose embeddings achieve only moderate performance on ontology-sensitive similarity judgments, with substantial degradation on hard negatives. Second, contrastive fine-tuning dramatically improves triplet discrimination, with hyperbolic objectives achieving near-perfect ranking accuracy. Third, these improvements transfer only weakly to downstream ontology engineering tasks such as taxonomy discovery~\cite{babaei2023llms4ol} and ontology alignment~\cite{hertling2023olala}. Together, these results reveal an optimization--generalization gap: \emph{embeddings can learn to discriminate AVA perturbations without acquiring robust, transferable representations of ontological structure and pattern-specific discrimination rather than general ontological reasoning}.

The contributions of this work are threefold: (1) we introduce AVA, a large-scale benchmark for evaluating ontology-aware semantic similarity using structured logic-sensitive perturbations; (2) we provide a comprehensive evaluation of modern embedding models and contrastive learning objectives, including Euclidean and hyperbolic formulations; and (3) we demonstrate that high contrastive discrimination accuracy does not necessarily imply transferable ontology understanding, highlighting important limitations of current embedding-based approaches for Semantic Web applications. Furthermore, we make the implementation publicly available to the research community at \url{https://github.com/sciknoworg/AVA}.

\begin{figure}
    \centering
    \includegraphics[width=\linewidth]{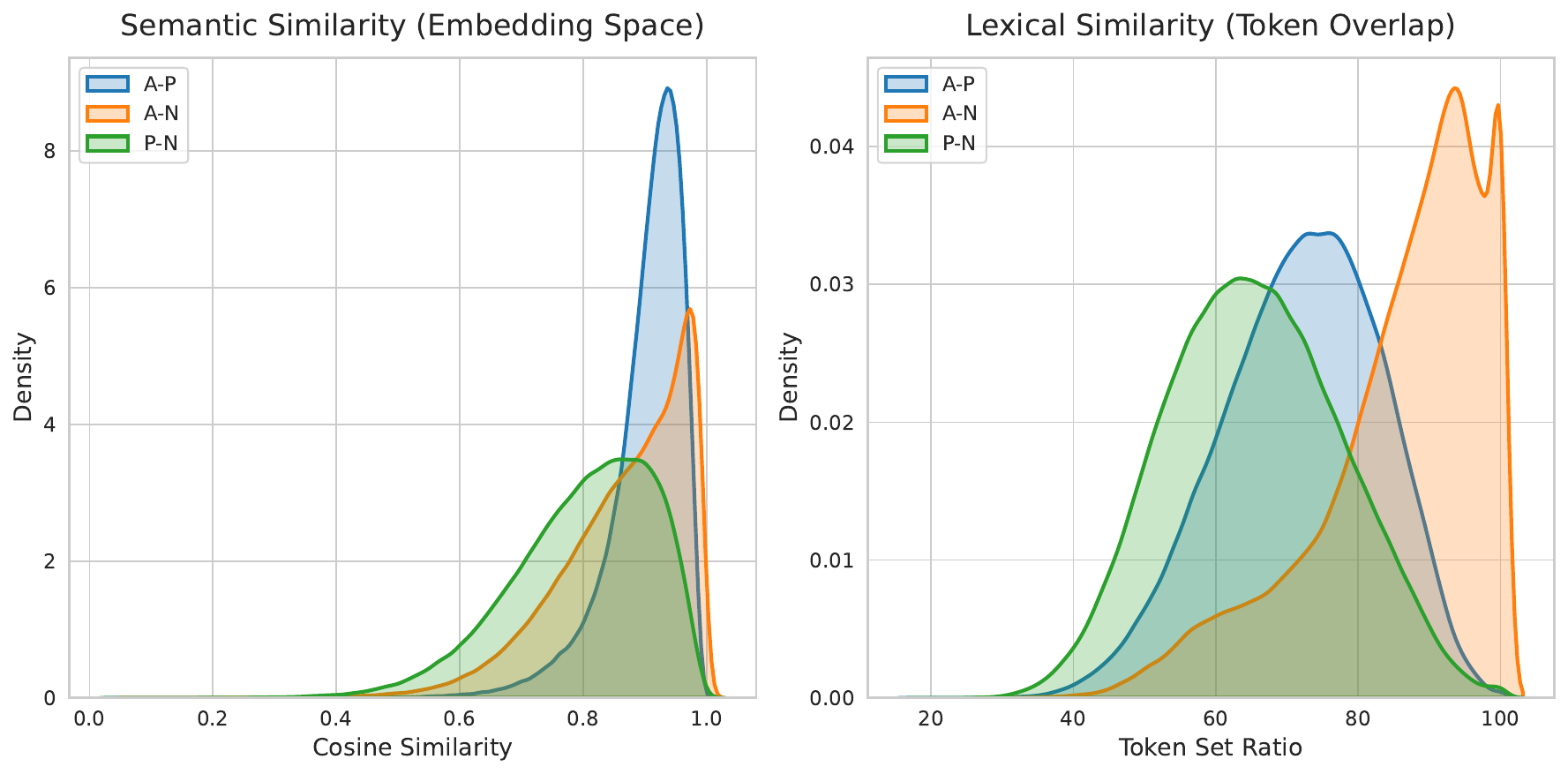}
    \caption{Dataset analysis:  semantic similarity (cosine similarity computed using MPNET-base) and lexical similarity (token-set ratio), distributions for anchor-positive (A-P),  anchor-negative (A-N), and positive-negative (P-N) pairs.}
    \label{fig:dataset_analysis}
\end{figure}
 
\begin{table}[t]
    \centering
    \caption{Examples of logic-sensitive hard negatives. }
    \label{tab:hard_negatives}
    \resizebox{0.49\textwidth}{!}{%
    \begin{tabular}{l}
    \hline
    \textbf{Anchor}: An online gaming account is a subclass of an \textcolor{blue}{online account}.   \\
    \textbf{Hard Negative}: An online gaming account is a subclass of an \textcolor{red}{agent}. \\
    \hline
     \textbf{Anchor}: Online chat accounts are defined as a subclass of \textcolor{blue}{online accounts}. \\
     \textbf{Hard Negative}: Online chat accounts are defined as a subclass of \textcolor{red}{online e-commerce accounts}. \\
     \hline
     \textbf{Anchor}: An online gaming account is a specific type of \textcolor{blue}{online accounts}. \\
     \textbf{Hard Negative}: An online gaming account is a specific type of \textcolor{red}{online chat account}.\\
     \hline
    \end{tabular}
    }
\end{table}

\begin{table}[t]
    \centering
    \caption{Dataset statistics.}
    \label{tab:dataset_stats}
    \resizebox{0.5\textwidth}{!}{%
    \begin{tabular}{lr}
        \hline
         & \textbf{Count} \\
         \hline
        \textit{Ontologies} &  163 \\
        \textit{BFS subgraphs} & 50,548 \\
        \textit{Synthesized samples (raw)} & 197,326 \\
        \textit{Removed (Positive-Negative pairs that are too similar)} & 4,154\\
        \textit{After cleaning \& de-duplication} & 171,007 \\
        \textit{Hard negatives (A-N similarity $\geq 90$)} & 77,932\\
        \textit{Mean sentence length (words)} & 10\\
        \hline
        \textit{Train (Hard Negatives)} & 153,211 (70,703) \\ 
        \textit{Test (Hard Negatives)} & 17,796 (7,229) \\ 
        \hline
    \end{tabular}
    }
\end{table}

\begin{figure}
    \centering
    \includegraphics[width=0.78\linewidth]{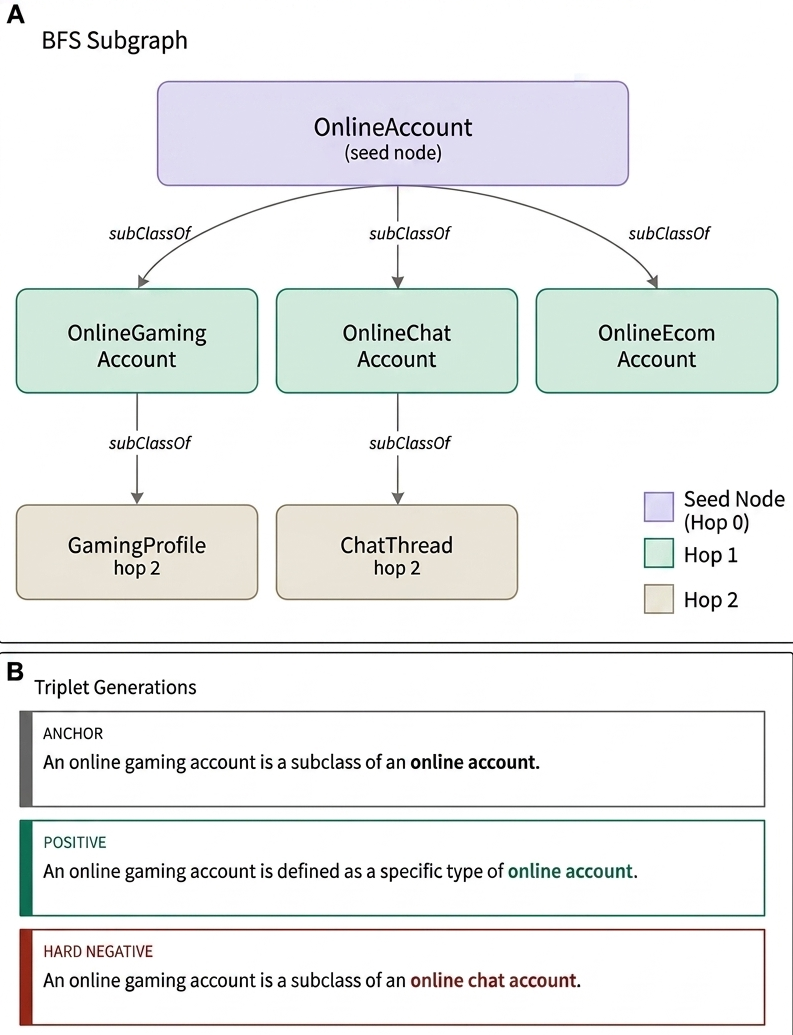}
    \caption{Overview of triplet generation. (A) A two-hop BFS subgraph extracted from an ontology. (B) Example contrastive triplet derived from the subgraph.}
    \label{fig:example_bfs_triplet}
\end{figure}

\section{AVA}

AVA is an evaluation framework for assessing whether embedding models capture ontology-level relational semantics beyond surface lexical similarity. It combines (1) ontology perturbations that generate logic-sensitive contrastive triplets and (2) contrastive objectives that evaluate ontology-aware discrimination and cross-ontology transfer using triplet loss, hyperbolic loss, and reinforcement learning techniques~\cite{christiano2017deep,lambert2022illustrating}.

\subsection{Structured Ontology Perturbations}
\noindent\textbf{Ontology Graph Extraction.} We used 163 ontologies from diverse domains, including biomedical (i.e., GO~\cite{gene-ontology}, OBI~\cite{bandrowski2016ontology}), geospatial, social, engineering, and schema ontologies. We accessed these collections via the OntoLearner library~\cite{giglou2026ontolearner}. For each ontology, we construct an undirected graph in which nodes represent OWL classes and object/datatype/annotation properties, and edges encode structural OWL/RDFS relations (i.e., \texttt{rdfs:subClassOf}, \texttt{owl:domain}, \texttt{owl:range}, etc). Existential restriction axioms of the form $C \sqsubseteq \exists R.D$ are reified as direct labeled edges between $C$ and $D$. To construct plausible hard negatives—such as sibling swaps—without exceeding the LLM context window, we extract subgraphs using a two-hop breadth-first search (BFS) seeded at every OWL class. Single-hop neighborhoods frequently lack sibling classes, whereas a two-hop radius reliably captures the necessary sibling and grandparent relationships. We retain subgraphs bounded between 3 and 20 nodes to exclude trivial or unwieldy structures, and deduplicate them by the MD5 fingerprint of their sorted node sets, resulting in 50,548 unique subgraphs.

\begin{figure}[htbp]
        \centering
        \begin{tcolorbox}[
            colback=gray!5!white,
            colframe=gray!75!black,
            title=\textbf{Contrastive Triplet Generation Prompt},
            fonttitle=\bfseries,
            arc=0.5mm,
            boxrule=0.4pt,
            left=6pt, right=6pt, top=6pt, bottom=6pt
        ]
        {\fontsize{6.3pt}{7pt}\selectfont
        \textbf{\textless task\textgreater}\\
        The following content represents a structural sub-graph extracted from a formal knowledge ontology, detailing entities, their definitions, and their semantic relationships (triples).\\
        \textbf{\textless/task\textgreater}
        
        \vspace{3pt}
        \textbf{\textless objective\textgreater}\\
        Synthesize a high-quality dataset of natural language sentence triplets (Anchor, Positive, Negative). This dataset will be utilized for contrastive learning to fine-tune an embedding model (e.g., a Sentence-Transformer) for semantic representation and ontology alignment.\\
        \textbf{\textless/objective\textgreater}
        
        \vspace{3pt}
        \textbf{\textless guidelines\textgreater}\\
        Guidelines for Triplet Generation:\\
        1. \textbf{Anchor}: Formulate a clear, unambiguous declarative sentence that reflects a true fact, definition, or exact relation directly asserted explicitly in the provided sub-graph.\\
        2. \textbf{Positive}: Construct a semantically equivalent paraphrase of the Anchor. This sentence must preserve the exact factual meaning but utilize morphological variations, synonyms, or alternative syntactic structures (e.g., active versus passive voice) to encourage robust semantic representation.\\
        3. \textbf{Negative}: Construct a 'hard negative' statement. This sentence must exhibit high lexical overlap with the Anchor (employing similar entities, relations, or vocabulary from the ontology domain) but assert a factually incorrect relationship, a contradictory definition, or pair disjoint entities. It must serve as a difficult distractor for the embedding model.\\
        \textbf{\textless/guidelines\textgreater}
        
        \vspace{3pt}
        \textbf{\textless output-format\textgreater}\\
        - Generate exactly 5 unique triplets. Output strictly as a valid JSON array\\
        - Output the result strictly as a valid JSON array of objects, with the exact keys: "anchor", "positive", and "negative".\\
        
        Example Output: 
\begin{verbatim}
[{"anchor": "...", "positive": "...", "negative": "..."}]
\end{verbatim}
        \textbf{\textless/output-format\textgreater}
        
        \vspace{3pt}
        \textbf{\textless input-subgraph\textgreater}\\
        \texttt{\{subgraph\} }\\
        \textbf{\textless/input-subgraph\textgreater}
        }
        \end{tcolorbox}
        \caption{LLM prompt used for generating contrastive triplets. \texttt{\{subgraph\}} is a placeholder for structured presentation of subgraphs.}
        \label{fig:triplet-prompt}
    
        

    

\end{figure}

\noindent\textbf{Logic-Sensitive Hard Negative Synthesis.} Each subgraph is converted into a structured natural-language prompt presenting its entities with labels and definitions alongside their RDF triples. We instruct a \texttt{Qwen3.5-35B-A3B} LLM~\cite{qwen3.5} to generate five contrastive triplets per subgraph under the following constraints (see prompt at \autoref{fig:triplet-prompt}): \textit{1) Anchor}, a declarative sentence expressing a fact directly asserted in the subgraph (hierarchy assertion, domain/range constraint, equivalence, or disjointness);  \textit{2) Positive}, a semantically equivalent paraphrase using morphological variation, synonymy, or syntactic restructuring (e.g., active to passive voice). The factual content must be preserved exactly; \textit{3) Hard negative}, a statement exhibiting high lexical overlap with the anchor but asserting an ontologically incorrect relationship, for instance, swapping a subclass for a sibling, inverting a property domain, or replacing an entity with a disjoint concept. This synthesis procedure is specifically designed to produce \emph{logic-sensitive} negatives rather than randomly sampled distractors. After post-processing the raw outputs, we obtained a total of 197,326 candidate (Anchor, Positive, Negative) triplets. Representative examples are shown in \autoref{tab:hard_negatives}, and \autoref{fig:example_bfs_triplet} shows how, using BFS subgraphs, triplets are generated.
 
\noindent\textbf{Post-Processing and Dataset Statistics.} We apply two filtering passes using token-set ratio similarity scores computed with \textit{RapidFuzz}. First, we flag samples where the anchor–negative similarity exceeds 90 as hard negatives and move them to a dedicated evaluation split, as they represent the most challenging cases. Second, samples where the positive–negative similarity exceeds 90 are discarded outright, as the two cannot be meaningfully distinguished (w.r.t \autoref{fig:dataset_analysis}, the A-P distributions). After de-duplication at the anchor level, the dataset contains 171,007 samples. We perform the train/test split at the ontology level: all samples from a given ontology are assigned exclusively to train or test, with the test set constructed from ontologies whose label sets share fewer than 100 labels with the training ontologies. Source and subgraph disjointness is verified programmatically to ensure that there is no leakage between test and train sets, except for a small number of subgraphs (less than 0.1\% of triplets, which is $\approx$ 148 triplets) rooted in upper-ontology classes shared across multiple imported ontologies. Dataset statistics are summarized in \autoref{tab:dataset_stats}. General NLP embeddings (specifically MPNET-base) cannot separate  anchor-positive from anchor-negatives in semantic space, yet hard negatives are deliberately  more lexically similar to the anchor than the positives  (see \autoref{fig:dataset_analysis}). A model that learns to  discriminate these triplets, therefore, cannot rely on lexical  overlap and must learn something about relational structure.

\subsection{Contrastive Learning}
\noindent\textbf{Triplet Loss.} We fine-tune using the standard  cosine triplet loss, which optimizes the margin between  anchor-positive and anchor-negative similarity scores. Given  embeddings $\mathcal{A}$, $\mathcal{P}$, and $\mathcal{N}$,  the loss is $\mathcal{L}_{\text{tri}} = \max(0,\,  \mathrm{sim}(\mathcal{A}, \mathcal{N}) -  \mathrm{sim}(\mathcal{A}, \mathcal{P}) + m)$, where $m$ is  a margin hyperparameter set to $0.3$.


\noindent\textbf{Hyperbolic Triplet Loss.} Ontological class hierarchies are inherently tree-structured, which Euclidean space represents poorly. We replace the Euclidean margin with a Poincar\'{e}-ball distance~\cite{nickel2017poincare}, computing $\mathcal{L}_{\text{hyp}} = \max(0,\, d_c(\mathcal{A}, \mathcal{P}) - d_c(\mathcal{A}, \mathcal{N}) + m)$, where $d_c$ is the hyperbolic distance at curvature $c = 0.3$. This encourages the model to reflect hierarchical structure in its embedding geometry.

\noindent\textbf{Embedding-Adapted DPO.} We adapt Direct  Preference Optimization~\cite{rafailov2023direct} to the  embedding setting by treating cosine similarity as an implicit  reward. A frozen reference encoder $f_{\text{ref}}$ acts as a  regularizer, and the loss penalizes the policy encoder $f_{\theta}$ whenever its similarity margin $\Delta_\theta = \mathrm{sim}_\theta(\mathcal{A}, \mathcal{P}) - \mathrm{sim}_\theta(\mathcal{A}, \mathcal{N})$ falls behind the reference margin $\Delta_{\text{ref}}$, with temperature $\beta = 0.5$.

\begin{table}[t]
\centering
\caption{Evaluation results.}
\label{tab:benchmark_results}
\resizebox{0.5\textwidth}{!}{%
\begin{tabular}{lccc}
\hline
\multirow{2}{*}{\textbf{Model}}&  \multicolumn{3}{c}{\textbf{Cross-Ontology}} \\
\cline{2-4}
 & \textbf{Triplet} &  \textbf{R@1} & \textbf{Hard Neg}  \\
\hline
MiniLM-L6 & 0.657 &  0.657 & 0.467 \\
MPNET-base & 0.636 &  0.636 & 0.427 \\
RoBERTa-large & 0.621   & 0.621 & 0.412 \\ 
\hline
Nomic-embed & 0.602  & 0.602 & 0.363 \\
Nomic-embed-MoE & 0.584   & 0.584 & 0.333 \\

\hline
E5-small & 0.553   & 0.553 & 0.354 \\
E5-base & 0.506  & 0.506 & 0.305 \\
E5-large & 0.542  & 0.542 & 0.338 \\
Multilingual-E5-large & 0.633  & 0.633 & 0.394 \\
\hline
GTE-small & 0.650 & 0.650 & 0.433 \\
GTE-base & 0.632  & 0.632 & 0.402 \\
GTE-large & 0.647 &  0.647 & 0.417 \\
\hline
BGE-small & 0.609 &   0.609 & 0.370 \\
BGE-base & 0.604 &  0.604 & 0.356 \\
BGE-large & 0.558 &   0.558 & 0.302 \\
\hline
Qwen3-Embedding-0.6B & 0.739  & 0.739 & 0.572 \\
Qwen3-Embedding-4B & 0.709  & 0.709 & 0.523 \\
Qwen3-Embedding-8B & 0.736  & 0.736 & 0.564 \\
\hline
Text-Embedding-3-small & 0.442   & 0.442 & 0.201 \\
Text-Embedding-3-large & 0.388 & 0.388 & 0.155 \\
Text-Embedding-ADA-002 & 0.499 & 0.499 & 0.238 \\
\hline
EmbeddingGemma-300m & 0.455   & 0.455 & 0.203 \\
\hline
Llama-Embed-Nemotron-8B & 0.217   & 0.217 & 0.135 \\
\hline
Linq-Embed-Mistral  & 0.634   & 0.634 & 0.411 \\
\hline
\textbf{MiniLM-L6 + Triplet Loss} & 0.975   & 0.975 & 0.955 \\
\textbf{MiniLM-L6 + Hyperbolic Loss} &  0.981   & 0.981 & 0.966 \\
\textbf{MiniLM-L6 + DPO Loss} & 0.913  & 0.913 & 0.885 \\
\hline
\textbf{MPNET-base + Triplet Loss} &   0.985  & 0.985 & 0.972 \\
\textbf{MPNET-base + Hyperbolic Loss} &  0.989  & 0.989 & 0.980 \\
\textbf{MPNET-base + DPO Loss} &   0.876   & 0.876 & 0.874 \\
\hline

\end{tabular}
}
\end{table}

\begin{table*}[th]
    \small
    \centering
    \caption{Taxonomy discovery results on MPNET-Base/MiniLM-L6 models before/after fine-tuning using the perpetuated dataset.}
    \label{tab:taxonomy_learning}
    \resizebox{\textwidth}{!}{%
    \begin{tabular}{lccc ccc ccc ccc ccc}
    \hline
    \multirow{2}{*}{\textbf{Model}} & \multicolumn{3}{c}{\textbf{GO}} & \multicolumn{3}{c}{\textbf{SchemaOrg}} & \multicolumn{3}{c}{\textbf{SWEET}} & \multicolumn{3}{c}{\textbf{OBI}} & \multicolumn{3}{c}{\textbf{PO}}\\
    \cline{2-16}
    & \textbf{Rec@1} & \textbf{Rec@5} & \textbf{Rec@10}  &  \textbf{Rec@1} & \textbf{Rec@5} & \textbf{Rec@10}  & 
                     \textbf{Rec@1} & \textbf{Rec@5} & \textbf{Rec@10}  &  \textbf{Rec@1} & \textbf{Rec@5} & \textbf{Rec@10} &
                     \textbf{Rec@1} & \textbf{Rec@5} & \textbf{Rec@10} \\
    \hline
    \textbf{MiniLM-L6} &\textbf{ 0.058} & \textbf{0.150} &\textbf{0.183} &0.084 &\textbf{0.251} &0.337 &\textbf{0.037} &\textbf{0.092} &\textbf{0.118} &\textbf{0.075} &\textbf{0.179} &\textbf{0.229} &0.122 &0.281 &0.327 \\ 
    \textbf{ $\;\;$+ Triplet loss} & 0.037 &0.090 &0.116 &0.078 &0.225 &0.307 &0.027 &0.071 &0.092 &0.059 &0.143 &0.180 &0.116 &0.250 &0.299 \\ 
    \textbf{  $\;\;$+ Hyperbolic loss} & 0.049 &0.135 &0.176 &0\textbf{.088} &0.248 &\textbf{0.349} &0.032 &0.081 &0.105 &\textbf{0.075} &0.171 &0.216 &0.133 &\textbf{0.286} &\textbf{0.333}\\
    \textbf{  $\;\;$+ DPO loss} &  0.013 &0.027 &0.033 &0.025 &0.062 &0.090 &0.007 &0.017 &0.021 &0.021 &0.044 &0.053 &0.025 &0.051 &0.062\\
    
    \hline
    \textbf{MPNET-base } &    \textbf{0.059} &    \textbf{0.148} &    \textbf{0.183} &    0.093 &    0.256 &    0.340 &    0.033 &    0.088 &    0.115 &    0.076 &    0.177 &    0.229 &    0.105 &    0.268 &    0.327 \\
    \textbf{ $\;\;$+ Triplet Loss} &   0.036 &    0.088 &    0.115 &    0.092 &    0.223 &    0.312 &    0.029 &    0.077 &    0.102 &    0.056 &    0.140 &    0.185 &    0.080 &    0.209 &    0.267 \\
    \textbf{  $\;\;$+ Hyperbolic Loss} &   0.051 &    0.128 &    0.166 &    \textbf{0.100} &    \textbf{0.262} &    \textbf{0.358} &    \textbf{0.036} &    \textbf{0.091} &    \textbf{0.117} &    \textbf{0.082} &    \textbf{0.181} &    \textbf{0.231} &    \textbf{0.126} &    \textbf{0.275} &    \textbf{0.329}  \\
    \textbf{  $\;\;$+ DPO Loss} &  0.007 &    0.013 &    0.016 &    0.015 &    0.042 &    0.058 &    0.004 &    0.008 &    0.010 &    0.010 &    0.021 &    0.028 &    0.017 &    0.037 &    0.048  \\
    \hline
    \end{tabular}
    }
\end{table*}

\begin{table*}[t]
    \small
    \centering
    \caption{Ontology alignment results on MPNET-Base/MiniLM-L6 models before/after fine-tuning using the perpetuated dataset.}
    \label{tab:ontology_alignment}
    \resizebox{1\textwidth}{!}{%
    \begin{tabular}{lccc ccc ccc ccc}
    \hline
    \multirow{2}{*}{\textbf{Model}} & \multicolumn{3}{c}{\textbf{ENVO-SWEET~\cite{karam2020matching}}} & \multicolumn{3}{c}{\textbf{Mouse-Human~\cite{dragisic2017experiences}}} & \multicolumn{3}{c}{\textbf{MaterialInformation-MatOnto~\cite{nas2023mse}}} & \multicolumn{3}{c}{\textbf{Yago-Wikidata~\cite{fallatah2020gold}}}\\
    \cline{2-13}
     & \textbf{Rec@1} & \textbf{Rec@5} & \textbf{Rec@10}  &  \textbf{Rec@1} & \textbf{Rec@5} & \textbf{Rec@10}  & 
                       \textbf{Rec@1} & \textbf{Rec@5} & \textbf{Rec@10} &  \textbf{Rec@1} & \textbf{Rec@5} & \textbf{Rec@10} \\
    \hline
    \textbf{MiniLM-L6} & 0.604 &0.775& 0.819& 0.852 &{0.934} &0.950& 0.325& 0.536 & 0.609&{0.905} & {0.964}& 0.974\\ 
    \textbf{ $\;\;$+ Triplet loss} & 0.598& 0.785 & 0.830 & 0.843 & 0.916 & 0.926 & 0.328 &\textbf{0.623}&\textbf{0.669}& 0.891 & 0.951& \textbf{0.980}\\
    \textbf{  $\;\;$+ Hyperbolic loss} &  \textbf{0.605} & \textbf{0.796} & \textbf{0.837}& \textbf{0.859} & 0.933 &\textbf{0.951} &\textbf{0.341}& 0.609 & 0.662& 0.901 & 0.951 & 0.974\\
    \textbf{  $\;\;$+ DPO loss} &  0.472 & 0.519 & 0.524& 0.685 & 0.715 & 0.724 & 0.119 & 0.162 & 0.166& 0.549 & 0.609 & 0.638\\
    \hline
    \textbf{MPNET-base } &   0.611 &    0.773 &    0.799 &    0.856 &    0.931 &    0.946 &    0.315 &    0.576 &    0.652 &    0.898 &    \textbf{0.974 }&    0.980 \\
    \textbf{ $\;\;$+ Triplet Loss} &   0.622 &    0.789 &    0.820 &    0.838 &    0.907 &    0.925 &    \textbf{0.358} &    \textbf{0.609} &    \textbf{0.725} &    0.905 &    0.964 &    0.974   \\
    \textbf{  $\;\;$+ Hyperbolic Loss} &   \textbf{0.626} &    \textbf{0.805} &    \textbf{0.834} &    \textbf{0.871} &    \textbf{0.937} &    \textbf{0.954} &    0.354 &    0.550 &    0.662 &    \textbf{0.931} &    0.967 &    \textbf{0.984}  \\
    \textbf{  $\;\;$+ DPO Loss} &   0.468 &    0.538 &    0.554 &    0.662 &    0.697 &    0.707 &    0.156 &    0.219 &    0.252 &    0.530 &    0.586 &    0.641\\
    \hline
    \end{tabular}
    }

\end{table*}

\section{Results}
Evaluation is performed under a cross-ontology setting where evaluation ontological samples are fully excluded from training. We use L2-normalized embeddings and cosine similarity $\mathrm{sim}(x,y)=\frac{x \cdot y}{|x||y|}$. Performance is measured via \textit{triplet accuracy}, defined as the proportion of cases where $\mathrm{sim}(\mathcal{A},\mathcal{P})>\mathrm{sim}(\mathcal{A},\mathcal{N})$ (ties counted as incorrect). We also report \textit{Hard Negative Accuracy} on samples where anchor–negative lexical similarity exceeds 90 (token-set similarity), measuring how often the model correctly ranks the positive above the hard negative ($sim (\mathcal{A}, \mathcal{P}) > sim (\mathcal{A}, \mathcal{N})$).  

\noindent\textbf{Performance of Pre-trained Embeddings.} The \autoref{tab:benchmark_results} summarizes the performance of more than 25 pre-trained embedding models. Overall, results reveal substantial limitations in current general-purpose embeddings when confronted with ontology-sensitive semantic distinctions. Among all evaluated models, \textit{Qwen3-Embedding-0.6B} achieves the strongest performance, reaching a triplet accuracy of $0.739$ and a hard negative accuracy of $0.572$. Larger variants of the same family show comparable results, suggesting that model scale alone does not guarantee improved ontological discrimination. Sentence embedding models such as \textit{MiniLM-L6}, \textit{MPNET-base}, \textit{GTE}, and \textit{multilingual-E5} achieve moderate performance, with triplet accuracies generally between $0.63$ and $0.66$. In contrast, several embedding models perform substantially worse; specifically, the OpenAI \textit{Text-Embedding-3-large} reaches only $0.388$ triplet accuracy, while \textit{Llama-Embed-Nemotron-8B} obtains $0.217$ triplet accuracy and only 0.135 hard negative accuracy. 

A consistent trend across all models is the large performance drop on hard negatives. Although some embeddings achieve reasonable overall triplet accuracy, they still struggle to distinguish ontology-consistent statements from highly similar contradictory statements. This suggests that many embeddings rely primarily on lexical and distributional similarity rather than explicit sensitivity to relational semantics such as subclass structure, domain/range constraints, or disjointness relations. These findings indicate that strong performance on standard semantic similarity or retrieval benchmarks does not necessarily translate into competence on ontology-level understanding or discrimination. 

\noindent\textbf{Effect of Contrastive Fine-Tuning.}  Fine-tuning dramatically improves triplet discrimination performance. Across both \textit{MiniLM-L6} and \textit{MPNET-base} (a widely used model in semantic web engineering tasks), all three optimization objectives substantially outperform their corresponding pre-trained baselines. For \textit{MPNET-base}, standard triplet loss increases triplet accuracy from $0.636$ to $0.985$, while hyperbolic triplet loss further improves performance to $0.989$. Hard negative accuracy exhibits a similar pattern, increasing from $0.427$ to $0.972$ and $0.980$, respectively. Comparable improvements are observed for \textit{MiniLM-L6}, where hyperbolic loss achieves the strongest overall results with $0.981$ triplet accuracy and $0.966$ hard negative accuracy. The hyperbolic objective consistently outperforms Euclidean triplet loss by a small but measurable margin. This result is consistent with prior works suggesting that hyperbolic geometry provides a more natural representation space for hierarchical structures~\cite{nickel2017poincare,dhingra2018embedding}. In contrast, the embedding-adapted DPO objective performs substantially worse than both triplet-based approaches, although it still improves considerably over the corresponding pre-trained models. This suggests the primary bottleneck may be reliance on a reference module whose representations do not encode fine-grained ontological distinctions. Consequently, preference optimization is likely constrained by the semantic limitations of the reference model, reducing its ability to learn transferable ontology-level representations.

Taken together, these results demonstrate that ontology-aware contrastive supervision enables embeddings to separate semantically valid statements from logic-sensitive negatives with near-perfect accuracy. A similar scenario is also observed in KEPLER~\cite{wang2021kepler}, where the fine-tuned model on the generated dataset performed very well. However, high performance alone might not establish that models have acquired transferable ontological understanding. Instead, the models likely learn to recognize perturbation-specific structural patterns rather than genuine symbolic semantics.

\noindent\textbf{Transfer to Downstream Ontology Engineering Tasks.} Despite the dramatic gains, downstream evaluation reveals a substantial optimization--generalization gap. The \autoref{tab:taxonomy_learning} reports taxonomy discovery task performance, where the aim is that, for a given set of ontological classes, the models should construct taxonomic pairs (parent, child) that form a subclass (or is-a) relation (experiments were performed with the OntoLearner~\cite{giglou2026ontolearner} library). Contrary to expectations, fine-tuning often degrades performance relative to the original models. Standard triplet loss consistently reduces recall across most ontology benchmarks. Hyperbolic loss partially mitigates this decline and occasionally yields modest improvements, particularly on SchemaOrg, OBI, SWEET, and Plant Ontology (PO), but gains remain relatively small compared to the near-perfect improvements on earlier stages. DPO fine-tuning produces severe degradation across all datasets. A similar pattern appears in ontology alignment (see \autoref{tab:ontology_alignment}), the task of finding equivalent classes between two different ontologies (experiments were conducted using the OntoAligner~\cite{babaei2025ontoaligner} library). Triplet and hyperbolic losses provide only modest improvements on several benchmarks, including ENVO--SWEET and MaterialInformation--MatOnto, while performance on other datasets remains largely unchanged. Hyperbolic loss achieves the strongest overall transfer performance, but improvements are typically measured in only a few percentage points. In contrast, DPO again substantially reduces alignment accuracy across all evaluation datasets.

The discrepancy between near-perfect triplet ranking accuracy and comparatively small downstream gains suggests that optimization does not necessarily induce robust ontology-level understanding. Instead, models may learn perturbation-specific decision boundaries that effectively distinguish the generated negatives without acquiring transferable representations of hierarchical and logical structure. These findings indicate that contrastive discrimination and ontology generalization should be treated as distinct evaluation objectives rather than interchangeable measures of semantic understanding.

\section{Discussion}
Because AVA's triplets are synthesized by a single LLM (Qwen3.5-35B-A3B) and the strongest pretrained encoder in our evaluation belongs to the same model family (Qwen3-Embedding), we cannot rule out that part of its advantage reflects shared stylistic or distributional artifacts between generator and encoder rather than genuinely superior ontological sensitivity. We treat this as a limitation of single-generator benchmark construction; incorporating multiple LLMs for triplet synthesis would increase lexical and stylistic diversity and further mitigate generator-specific artifacts, and we leave this, together with broader cross-generator validation, to future work.

We also note that, as commonly observed for retrieval modules in RAG pipelines, taxonomy discovery performance degrades when candidates exhibit high semantic overlap. Analysis in \cite{giglou2026ontolearner} indicates substantial semantic overlap among candidate classes across the ontologies used here, suggesting that the persistently low absolute recall observed for all models partly reflects this intrinsic task difficulty, which ontology-aware fine-tuning only partially mitigates.

Furthermore, the AVA specifically evaluates ontology discrimination rather than logical reasoning. It measures whether a model can distinguish between two lexically similar statements according to their ontological consistency, but it does not assess inference or entailment. Consequently, performance on the AVA should not be interpreted as evidence of logical reasoning ability. A separate evaluation would be required to test reasoning, for example, by assessing whether a model can infer $A \sqsubseteq C$ from $A \sqsubseteq B$ and $B \sqsubseteq C$ when $A \sqsubseteq C$ is not explicitly asserted. Such a setting would directly evaluate the model's ability to derive new knowledge from formal ontology axioms rather than merely discriminate between ontology-consistent and ontology-violating statements.

\section{Conclusion}
In conclusion, while embedding-based models can perform well on similarity and retrieval tasks, this does not necessarily imply that they can reliably discriminate between ontology-consistent and ontology-violating statements. In ontology engineering, such models may capture statistical patterns in the data while failing to reflect formal logical constraints. It therefore remains unclear whether embedding optimization alone is sufficient for reliable ontology-level discrimination, or whether explicit logical mechanisms are needed to adequately capture such constraints.

\begin{acks}
This work is supported by the  \href{https://www.nfdi4datascience.de/}{NFDI4DataScience initiative} (DFG, German Research Foundation, Grant ID: 460234259).
\end{acks}

\section*{GenAI Disclosure}
In preparing this manuscript, generative AI tools, specifically ChatGPT and Gemini, were used solely for grammar checking, spelling checks, and readability of some sentences.  All suggested changes were carefully reviewed and adapted by the authors to ensure accuracy and appropriateness. The scientific content, research design, analysis, and conclusions were developed and verified exclusively by the authors without AI involvement. 

\bibliographystyle{ACM-Reference-Format}
\bibliography{sample-base}


\end{document}